%% file: main.tex
\documentclass[conference]{IEEEtran}
\usepackage[noadjust]{cite}
\usepackage{graphics} 
\usepackage{xcolor}
\usepackage{epsfig} 
\usepackage{epstopdf}
\usepackage{amsmath} 
\usepackage{amssymb}  
\usepackage{stmaryrd}
\usepackage{transparent}
\usepackage{import}

\usepackage{siunitx}
\usepackage{subfig}
\usepackage[font=footnotesize,labelfont=bf]{caption}
\usepackage{url}
\usepackage{hyperref}

\usepackage{tikz}
\usepackage{siunitx}

\usetikzlibrary{shapes,arrows,positioning,calc}

\input{macros}

\newif\ifrevision
\revisionfalse
\ifrevision
  \newcommand{\rev}[1]{\textcolor{red}{#1}}
  
\else
  \newcommand{\rev}[1]{#1}
  
\fi

\begin{document}

%

\title{Learning dynamics for improving control of overactuated flying systems}
\author{Weixuan Zhang, Maximilian Brunner, Lionel Ott, Mina Kamel, Roland Siegwart, and Juan Nieto}

\maketitle

\begin{abstract}
Overactuated omnidirectional flying vehicles are capable of generating force and torque in any direction, which is important for applications such as contact-based industrial inspection.
This comes at the price of an increase in model complexity.
These vehicles usually have non-negligible, repetitive dynamics that are hard to model, such as the aerodynamic interference between the propellers.
This makes it difficult for high-performance trajectory tracking using a model-based controller. 
This paper presents an approach that combines a data-driven and a first-principle model for the system actuation and uses it to improve the controller.
In a first step, the first-principle model errors are learned offline using a Gaussian Process (GP) regressor. At runtime, the first-principle model and the GP regressor are used jointly to obtain control commands. This is formulated as an optimization problem, which avoids ambiguous solutions present in a standard inverse model in overactuated systems, by only using forward models.
The approach is validated using a tilt-arm overactuated omnidirectional flying vehicle performing attitude trajectory tracking. 
The results show that with our proposed method, the attitude trajectory error is reduced by 32\% on average as compared to a nominal PID controller. 

\end{abstract}


%
\IEEEpeerreviewmaketitle

\input{01_Intro}
\input{02_Related_work}
\input{03_Platform}
\input{04_Modeling}
\input{05_Approach}
\input{06_Experiments}
\input{07_Conclusion}

\bibliographystyle{IEEEtran}
\bibliography{IEEEabrv,bibliography}
%



\end{document}

%% file: macros.tex
\newcommand{\mvec}[1]{\boldsymbol{#1}}

\newcommand{\position}{\boldsymbol{p}}

\newcommand{\naturalFreqPos}{\omega_{n,p}}
\newcommand{\dampingRatioPos}{\zeta_p}

\newcommand{\naturalFreqAtt}{\omega_{n,a}}
\newcommand{\dampingRatioAtt}{\zeta_a}
\newcommand{\posRef}{\mvec{p}_{\mathrm{des}}}
\newcommand{\posErr}{\mvec{e}_p}
\newcommand{\posRateErr}{\dot{\mvec{e}}_p}
\newcommand{\posErrDD}{\ddot{\mvec{e}}_p}
\newcommand{\angErr}{\mvec{e}_\rotMat}
\newcommand{\angRateErr}{\mvec{e}_\omega}
\newcommand{\angAccErr}{\dot{\mvec{e}}_\omega}

\newcommand{\discrepancyFunc}{\eta}

\newcommand{\forceVec}{\mvec{F}}
\newcommand{\forceVecCmd}{\mvec{F}_{\mathrm{cmd}}}
\newcommand{\forceVecDes}{\mvec{F}_{\mathrm{des}}}
\newcommand{\wrenchVec}{\mvec{W}}

\newcommand{\wrenchVecCmd}{\mvec{W}_{\mathrm{cmd}}}

\newcommand{\wrenchVecDes}{\mvec{W}_{\mathrm{des}}}
\newcommand{\torqueVec}{\mvec{M}}
\newcommand{\torqueVecCmd}{\mvec{M}_{\mathrm{cmd}}}
\newcommand{\torqueVecDes}{\mvec{M}_{\mathrm{des}}}
\newcommand{\torqueVecMeas}{\mvec{M}_{\mathrm{meas}}}
\newcommand{\torqueVecPred}{\mvec{M}_{\mathrm{pred}}}
\newcommand{\allocMapping}{n}
\newcommand{\R}{\mathbb{R}}
\newcommand{\noise}{\epsilon}
\newcommand{\noiseVar}{\sigma^2}
\newcommand{\GPobservation}{z}
\newcommand{\GPobservationStacked}{\mvec{Z}}
\newcommand{\GPinput}{\mvec{\xi}}
\newcommand{\GPinputStacked}{X}
\newcommand{\GPtestInput}{\mathrm{\mvec{\xi}}^*}
\newcommand{\approximateActuationFuncToTorque}{h}

\newcommand{\controlInputsCmd}{\mvec{u}_{\mathrm{cmd}}}
\newcommand{\trueActuationFunc}{h^*}

\newcommand{\GPmean}[1]{m(#1)}
\newcommand{\GPcovariance}[1]{k(#1)}
\newcommand{\GPpredictionMean}[1]{\mu(#1)}

\newcommand{\GPpredictionMeanSingle}[2]{\mu_{#1}(#2)}
\newcommand{\GPpredictionVarSingle}[2]{\sigma^2_{#1}(#2)}
\newcommand{\unknownDynamicsFunc}[1]{g(#1)}
\newcommand{\unknownDynamicsFuncSingle}[2]{g_{#1}(#2)}

\newcommand{\bodyRates}{\mvec{\omega}_{BE}}

\newcommand{\cSys}[1]{^{#1}} 

\newcommand{\rotMat}{\ensuremath{\mathbf{R}}}

\newcommand{\dWrenchVecDes}{\Delta\wrenchVec}
\newcommand{\dWrenchVecDesPrev}{\Delta \mvec{W}_{\mathrm{prev}}}
\newcommand{\testPointOne}{\GPinput_p}
\newcommand{\testPointTwo}{\GPinput_q}

\newcommand{\norm}[1]{\left\| #1 \right\|}

\newcommand{\figref}[1]{Fig.~\ref{#1}}
\newcommand{\state}{\boldsymbol{x}}
\newcommand{\virtualInput}{\boldsymbol{v}}

%% file: 01_Intro.tex
\section{Introduction}\label{sec:introduction}
Omnidirectional flying vehicles (\cite{ryll2012modeling, brescianini2018omni,bodie2019omnidirectional, park2018odar, staub2018towards, kamel2018voliro}) are suitable for industrial inspection or interaction applications due to their decoupled translational and rotational dynamics (e.g. being able to hover at arbitrary attitudes) and the ability to exert forces and torques in arbitrary directions.

These vehicles are overactuated, resulting in a mechanical redundancy that provides the system with the necessary control authority to achieve omnidirectionality and increases their maneuverability and force/torque margin.
However, the increased mechanical complexity also poses some challenges.
Unlike standard quadrotors or hexacopters, for which a simple and accurate actuation model is available for low speed maneuvers \cite{mahony2012multirotor}, only an inaccurate and simplified actuation model can be obtained for omnidirectional vehicles from first principles. 
The actuation model is the mapping from the individual actuator commands to the total wrench, i.e., force and torque.
To achieve omnidirectionality, vehicle designs often cause complex aerodynamic effects that are hard-to-model and non-negligible, such as the aerodynamic interference between the rotors.
This is disadvantageous from a control perspective. A standard approach for overactuated flying vehicles is that a controller first computes a desired wrench, which, based on the inverse of the actuation model, is transformed into individual actuator commands. 
Due to unmodeled effects, this typically results in a different actual wrench than the intended one. This leads to degraded control performance and reduces the ability to accurately track desired trajectories.

\begin{figure}
    \centering
    \resizebox{0.49\columnwidth}{!}{\includegraphics{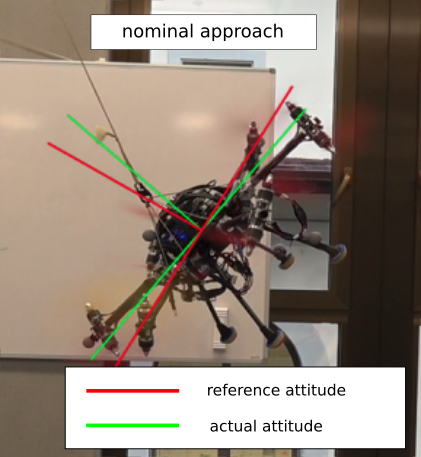}   }
    \hfill
    \resizebox{0.49\columnwidth}{!}{\includegraphics{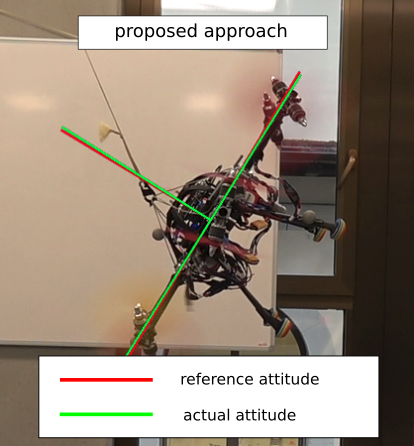}   }
    \caption{Attitude tracking performance of an overactuated flying vehicle: (left) without model learning, and (right) using our proposed learning based approach. \rev{For perfect tracking the green frame should coincide with the red frame.} }
    \label{pic:volirox}
\end{figure}
To solve the problem mentioned above, adding only integral actions in the feedback controller is often not enough, as they typically decrease the damping or stability of the system and are not able to react to fast-changing modeling errors. 
\rev{Another solution is to treat the unmodelled dynamics as disturbances \cite{hentzen2019disturbance} \cite{yuksel2014nonlinear} and employ a disturbance observer. While being computationally efficient, this reactive strategy introduces delay in tracking.}
One \rev{alternative} solution is to identify these mismatched dynamics and compute additional feedforward signals in an attempt to cancel out the effect of modeling errors. One way to model this mismatch is using data-driven machine learning models. Such models have recently been shown to be viable due to their ability to learn complex and nonlinear dynamic functions \cite{hwangbo2019learning}. 
Directly learning an inverse mapping of the actuation, that is, a mapping from the wrench to the individual actuator commands, is a commonly used method. However, this approach is problematic for an overactuated system due to its one-to-many mapping between wrench and actuator inputs. \rev{The multi-modal nature of this mapping has the risk of producing invalid results when averaging over multiple distinct modes with standard inverse model approaches.} 


This paper presents an approach aiming to improve the control of overactuated flying systems. First, the actuation modeling error is learned offline using a Gaussian process (GP) regressor. To overcome the challenge arising due to the above-mentioned one-to-many mapping, we formulate the task of finding a suitable wrench command as an iterative optimization problem. \rev{We obtain} a wrench by optimizing the commanded wrench using the analytical forward model and its learned error model. This yields in expectation the desired behaviour of the system, necessitating no inverse models.


This paper makes the following contributions:
\begin{itemize}
    \item Proposal of a \rev{GP-based} model to capture the model-plant mismatch common among overactuated omnidirectional flying vehicles.
    \item Introduction of an optimization-based method using a first-principles model and a learned GP error model to select a signal correcting for the model-plant mismatch, necessitating only forward models.
    \item Experimental validation of the proposed approach on a tilt-arm overactuated flying vehicle called VoliroX \cite{bodie2019omnidirectional} (\figref{pic:volirox}), with a reduction of the attitude tracking error of 32\% on average. 
\end{itemize}

%% file: 02_Related_work.tex
\section{Related work}\label{sec:related_work}
Optimal control such as the model predictive controller is shown to work well for standard multicopters \cite{kamel2017linear}. However, it is shown in \cite{brunner2019} that the model predictive controller has comparable performance to a standard PD controller in case of insufficient model knowledge.

In this section, we focus mainly on the review of approaches that learn the nonlinear, non-negligible model-plant-mismatch, which is then used to improve control performance. A broader review of model learning for control can be found in \cite{nguyen2011model}. 
We categorize the related work by the types of model being used: forward models and inverse models. 
A forward model could be a static mapping (e.g. an actuator mapping from individual actuator commands to the resulting wrench in case of neglecting its dynamics) or a dynamics model (the mapping from the state and the control inputs to the derivative of the state), which are uniquely defined. The inverse model is the inverse of the forward model. 

\subsubsection{Inverse model approach}
\rev{The inverse model approach aims to find a mapping that maps from the robot state to a feedforward control command.}
Several approaches \cite{chowdhary2014bayesian, nguyen2009local, meier2016towards, beckers2017stable, helwa2019provably} use a GP regression approach to model the inverse dynamics. The implied assumption of these approaches is that there is a unique mapping between the input and output data so that a direct regression is well-defined.  Unfortunately, for an overactuated platform this is not the case.

\subsubsection{Forward model approach}
Forward model learning has been combined with iterative linear quadratic Gaussian algorithm \cite{mitrovic2010adaptive}, or model predictive control \cite{kabzan2019learning, ostafew2014learning, cao2017gaussian}. These methods use rollouts of the learned dynamics in an optimization-based framework to find a sequence of control actions in real-time. 
In \cite{kabzan2019learning, cao2017gaussian, ostafew2014learning}, the model-plant mismatches are modeled using GPs and embedded into a model predictive controller. These methods are computationally expensive and thus they typically use approximation techniques to propagate the state distribution in the rollout. They experimentally validated these approaches (\cite{kabzan2019learning, ostafew2014learning}) on autonomous mobile vehicles.
In \cite{cao2017gaussian} the authors demonstrated the proposed approach in the simulation of a quadrocopter. One of the aims of our work is to validate the proposed approach on a real system. 

Model-based reinforcement learning control may also be employed with a learned forward model. 
Among these approaches, the control policy is either found through experiments \cite{deisenroth2011pilco}, which needs to make sure that the system does not damage the robots, or through simulation \cite{hwangbo2019learning,saveriano2017data, ko2007gaussian}. 
More specifically, the complete actuator model \cite{hwangbo2019learning} or the residual dynamics (\cite{saveriano2017data, ko2007gaussian}) are first learned using either a neural network or a GP and then embedded into the simulation. A high-performance control policy is then obtained in the simulation through reinforcement learning.
These approaches typically require a large amount of training data.

Another approach is given a desired state, the control inputs are solved in an equation that includes the forward model. In \cite{spitzer2019inverting}, in order to improve the tracking performance of a quadrocopter, an acceleration error model is learned using the linear regression. It is then embedded into the acceleration model equation, in which the body angular acceleration command is solved for in real-time as the control inputs. It does not take uncertainty into consideration.






%% file: 03_Platform.tex

\section{Platform description}
VoliroX (\figref{pic:volirox}) is an overactuated omnidirectional flying vehicle with six tiltable arms in a hexagonal arrangement. A coaxial rotor configuration is rigidly attached to the end of each arm. The propellers spin in the opposite direction, which helps to reduce the gyroscopic effects introduced by the rotation of the arm and thus the tilt motor efforts. 
The rotation of each arm can be actively controlled by a servo motor, which results in a total of 18 actuators. 
This setup allows for force- and pose omnidirectionality for almost all configurations and an improved hover efficiency when compared to fixed tilt omnidirectional vehicles.

%% file: 04_Modeling.tex
\section{Modeling}\label{sec:background}
This section describes the dynamics of the flying vehicle and presents its actuation model.

\subsection{Dynamic model}\label{sec:dyn}
The position of the vehicle's center of mass with respect to a point fixed in the inertial coordinate system is denoted as $\position \in \R^3$. Boldface symbols are used throughout the paper to denote vectors. Two types of coordinate systems are used for the modeling: an inertial coordinate system $E$, a body-fixed coordinate system $B$.
A vector expressed in a specific coordinate system is indicated by a superscript, for example $\position \cSys{E}$ expresses $\position$ in coordinate system $E$.
The vehicle rotates at an angular velocity $\bodyRates$. The subscript $BE$ in $\bodyRates$ denotes the relative velocity of coordinate system $B$ with respect to $E$.

Using a simplified physical model, the dynamics model can be written as follows \cite{bodie2019omnidirectional}:	
\begin{equation}\label{eq:sys_dyn}
    \dot{\state} = f(\state) + \virtualInput
\end{equation}
where $\state = [\dot{\position} \cSys{E}, \bodyRates\cSys{B}] $ and $\virtualInput =[\forceVec\cSys{E} / m, (\boldsymbol{I}_B^B)^{-1} \torqueVec\cSys{B}]$ with $m$ being the vehicle mass and $\boldsymbol{I}_B^B$ being the vehicle inertia matrix.   
The translational drag forces and rotational drag torques are neglected, as it is assumed that the vehicle travels at low translational and angular velocities.
We assume no model-plant mismatches in \eqref{eq:sys_dyn} as the rigid body dynamics are typically known and the parameters such as mass or inertia can be precisely obtained through CAD model.

\subsection{Approximate actuation model}
The design of the VoliroX leads to complicated aerodynamic effects (e.g. coaxial rotor configurations and interaction between the adjacent propeller flows), which are hard to model using  first-principles. Therefore, the following assumptions are made to approximate the actuation model:
\rev{1).} It is assumed that the force and torque produced by a stationary propeller are proportional to its angular velocity squared, independent of the vehicle's translational and angular velocity. The resulting force and torque vectors are also assumed to be parallel with each other and perpendicular to the rotor disk. \rev{2).} Every rotor has the same torque and force coefficient. \rev{3).} The rotors do not interfere with each other, therefore the total force and torque produced by the rotors are the sum of individual force and torque vectors. \rev{4)} Body drag forces and torques are neglected due to the low translational and angular velocity of the flying vehicle.
\rev{5)}. It is assumed that the actuators have no temporal dynamics, that is, the past wrench commands do not affect the current wrench. 

Based on the above assumptions, a nonlinear approximate mapping $\approximateActuationFuncToTorque(\cdot): \R^{18} \rightarrow \R^6 $  between the individual actuator commands and the total wrench command $\wrenchVecCmd \cSys{B} = [\forceVecCmd\cSys{B}, \torqueVecCmd\cSys{B}]$ expressed in the body coordinate system $B$ may be established \cite{bodie2019omnidirectional}. That is
\begin{equation} \label{eq:actuation_model}
  \wrenchVecCmd\cSys{B} = \approximateActuationFuncToTorque(\controlInputsCmd)
\end{equation}
where $\controlInputsCmd \in \R^{18}$ contains the twelve propeller thrusts and the six tilt arm angles. For the sake of brevity, the mapping $\approximateActuationFuncToTorque(\cdot)$ is not described in detail as it does not affect the subsequent derivation. 

This approximate modeling of the actuation introduces modeling errors, this results in a $\wrenchVec \cSys{B}$ which differs from the intended $\wrenchVecCmd\cSys{B}$:
\begin{equation} \label{eq:torque_definition}
  \wrenchVec \cSys{B} = \trueActuationFunc(\controlInputsCmd) = \approximateActuationFuncToTorque(\controlInputsCmd) + \discrepancyFunc(\controlInputsCmd)
\end{equation}
where $\trueActuationFunc(\cdot)$ is the true unknown actuation function and $\discrepancyFunc(\cdot)$ is an additive nonlinear modeling error function.

Note that this additive error modeling implies that $\discrepancyFunc(\cdot)$ and $\approximateActuationFuncToTorque (\cdot)$ do not need to have the same parametric structure, which allows for more flexibility in the modeling.
On the other hand, misidentification of $\discrepancyFunc(\cdot)$ can result in a physically implausible prediction of the generated wrench.

In addition, the functions $h(\cdot)$, $\trueActuationFunc(\cdot)$, and $\discrepancyFunc(\cdot)$ are surjective, that is, for each achievable wrench there are possibly more than one set of feasible inputs.



%% file: 05_Approach.tex
\section{Approach}\label{sec:method}


The problem we are addressing in this report is formulated as follows: Given the system dynamics \eqref{eq:sys_dyn}, the actuation model \eqref{eq:actuation_model}, the unknown modeling error $\discrepancyFunc(\controlInputsCmd)$, and a given desired wrench $\wrenchVecDes \cSys{B}$  which is the output of a given model-based nominal controller, find the actuator commands $\controlInputsCmd$ such that the $\wrenchVecDes \cSys{B}$ is achieved. 
We distinguish between $\wrenchVecDes \cSys{B}$, which is the output of the controller, and $\wrenchVecCmd \cSys{B}$, which is the input to the allocation.

\begin{figure*}[ht!]
  \centering  
  \resizebox{0.8\textwidth}{!}{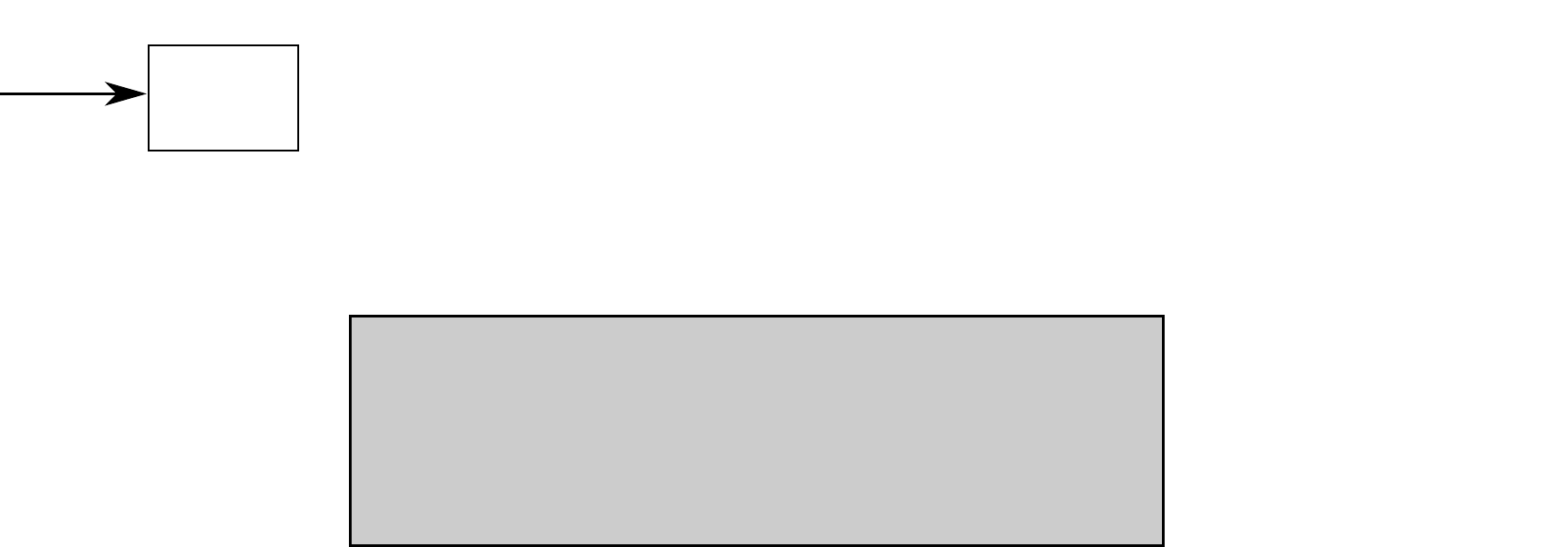}
  \caption{A block diagram of the proposed solution. The model-plant mismatch is modeled using GPs and regressed offline using flight data. The model is then fed into an optimization to find a compensation signal $\dWrenchVecDes$ so that the generated wrench $\wrenchVec\cSys{B}$ is equal to $\wrenchVecDes\cSys{B}$.} 
  \label{pic:control strategy}
\end{figure*}

The proposed solution aims to find a compensation signal \rev{at} the wrench level which, when applied to the nominal controller's command, results in the desired wrench being executed by the system. \rev{This reduces the problem to six dimensions since we do not search compensation signals for individual actuator commands $\controlInputsCmd$, as they are typically computed on a low-level flight computer that is computationally limited. Furthermore, more training data would be needed for a higher dimensional input space.}
The approach is summarized as follows and illustrated in \figref{pic:control strategy}: the nominal controller computes a desired wrench $\wrenchVecDes \cSys{B}$ according to the reference trajectory and the state estimate. 
The discrepancy between the model and the plant is modeled using GPs. 
The regressed model is input into an optimization framework to obtain an additive compensation signal to the desired wrench, which in case of high prediction uncertainty is filtered towards zero. 
The resulting command wrench $\wrenchVecCmd \cSys{B}$ is then sent to the control allocation, which outputs individual actuator commands $\controlInputsCmd$. 


This approach has the advantage that the GP and optimization block is modular and therefore can be added to the nominal controller.
Furthermore, an important point in our proposed solution is that the uncertainty information provided by the GP is used to prevent uncertain predictions from destabilizing the system. This is a key for safely deploying the proposed controller in a physical system.

\subsection{Nominal controller and control allocation}

Let $\posErr$ denote the position error of the vehicle (i.e, the difference between the actual position $\position$ and the desired position $\posRef$) and $\posRateErr$ its time-derivative.
Let $D$ denote the reference body frame and define the attitude error terms as follows:
\begin{align}
 \angErr &= \frac{1}{2} (\rotMat\cSys{DB} -\rotMat\cSys{BD})^\vee \label{eq:error_def_0} \\
 \angRateErr & = \mvec{\omega}_{BD}^{B} \label{eq:error_def_1}
\end{align}
where $\rotMat\cSys{DB}$ is the coordinate transformation matrix from the coordinate system $D$ to $B$. The mapping $\vee$ maps a skew-symmetric matrix to its $\R^3$ space.


Similar to the position control in \cite{zhang2016controllable}, the desired wrench $\wrenchVecDes \cSys{B} = [\forceVecDes\cSys{B}, \torqueVecDes\cSys{B}]$ is calculated such that
\begin{align}
	\posErrDD\cSys{E} &= - 2\dampingRatioPos\naturalFreqPos \posRateErr\cSys{E} - \naturalFreqPos^2  \posErr\cSys{E} \\
	\angAccErr\cSys{B} &= - 2\dampingRatioAtt\naturalFreqAtt \angRateErr\cSys{B} -\naturalFreqAtt^2 \angErr\cSys{B} \label{eq:attitude_control}
\end{align}
where ($\dampingRatioPos$, $\naturalFreqPos$) and ($\dampingRatioAtt$, $\naturalFreqAtt$) are a set of damping ratios and natural frequencies for the translational dynamics and attitude dynamics, respectively.

From \eqref{eq:error_def_0} and \eqref{eq:error_def_1} we can see that to first order, $\angRateErr$ is the time-derivative of $\angErr$. This means that if the desired wrench is tracked perfectly, the translation deviation $\posErr$ and the rotational deviation $\angErr$ will behave like a damped second order system. This makes the parameter tuning more intuitive.

Recall that the actuator dynamics are neglected, thus the wrench and the individual actuators are a static mapping. The wrench command then gets allocated through a chosen mapping $\allocMapping(\cdot) : \R^6 \rightarrow \R^{18}$ \cite{bodie2019omnidirectional} and a saturation function $\mathrm{sat}(\cdot): \R^{18} \rightarrow \R^{18}$ in the inner loop
\begin{equation}\label{eq:pseudo_inverse}
\controlInputsCmd =  \mathrm{sat}(\allocMapping(\wrenchVecCmd \cSys{B}))
\end{equation}
Again, we omit the details of the allocation for brevity. 
This allocation is essentially a pseudo-inverse mapping of $\approximateActuationFuncToTorque(\cdot)$. 
The mapping $\allocMapping(\cdot)$ is injective, that is, each set of actuation has at most one corresponding wrench. 


\subsection{Learning model-plant mismatch}
We first derive the model-plant mismatch as a function of the wrench command. 
Substituting \eqref{eq:actuation_model} and \eqref{eq:pseudo_inverse} into \eqref{eq:torque_definition} yields
\begin{align}
 \wrenchVec \cSys{B} 
&= \approximateActuationFuncToTorque(\controlInputsCmd) + \discrepancyFunc( \mathrm{sat}(\allocMapping(\wrenchVecCmd \cSys{B}))) \\
&= \wrenchVecCmd\cSys{B} + \unknownDynamicsFunc{\wrenchVecCmd\cSys{B}} \label{eq:torque_definition2},
\end{align}
where $\unknownDynamicsFunc{\cdot} : \R^6 \rightarrow \R^6$ is the composition of the injective mapping $\allocMapping(\cdot)$, $\mathrm{sat}(\cdot)$ and the surjective mapping $\discrepancyFunc(\cdot)$.
It may therefore be a surjective mapping.
This means that for each $\wrenchVec \cSys{B}$ there may be more than one set of possible $\wrenchVecCmd\cSys{B}$ that achieves it. This leads to the problem mentioned in Section \ref{sec:introduction}: an inverse mapping from $\wrenchVec \cSys{B}$ to $\wrenchVecCmd\cSys{B}$ is a one-to-many mapping and may not directly be learned.

The unknown error function $\unknownDynamicsFunc{\cdot}$ is modeled as six independent GPs. We make the assumption that the outputs of these GPs are uncorrelated. 
A GP is a collection of random variables, any finite number of which have a joint Gaussian distribution \cite{rasmussen2003gaussian}, and can be viewed as a distribution over functions.
It is characterized by a mean function $\GPmean{\cdot}$ and a covariance function $\GPcovariance{\cdot, \cdot}$, which is a positive-definite kernel function parametrized by a set of hyperparameters.

A \rev{GP-based} representation was chosen because it handles stochasticity (e.g. measurement noise) naturally, its nonparametric property offers flexibility in the modeling, and gives us information about the uncertainty of predictions made. 
On the other hand, it is known that the GP prediction does not scale well with data (it has a computational complexity of $\mathcal{O}(N^3)$, where $N$ is the number of data samples). 
Reducing the computational complexity is beyond the scope of this work. We are currently exploring the use of local GPs to handle the large state space.


We may apply Gaussian process regression to approximate the error function $\unknownDynamicsFuncSingle{l}{\wrenchVecCmd\cSys{B}}$ according to \eqref{eq:torque_definition2}, with the subscript $l$ denoting the $l$-th entry of the function $\unknownDynamicsFunc{\cdot}$: 
\begin{equation} \label{eq:torque_regression_model}
  \GPobservation =\unknownDynamicsFuncSingle{l}{\GPinput} + \noise
\end{equation}
where $\GPobservation \in \R$ denotes the observation of the $l$-th entry of $\wrenchVec \cSys{B} - \wrenchVecCmd \cSys{B}$, $\noise \in \R$ is the measurement noise with Gaussian distribution $\mathcal{N}(0,\noiseVar)$ \rev{with $\noiseVar$ being the variance}, and $\GPinput \in \R^6$ denotes $\wrenchVecCmd\cSys{B}$ for brevity for the remaining of this section.

Let $(\GPinput_i, \GPobservation_{i})$ denote one observed data point and \rev{$\GPinputStacked \in \R^{(N+1)\times6}, \GPobservationStacked \in \R^{N+1}$} denote the stacked version of $(\GPinput_{0}, ..., \GPinput_{N})$, $(\GPobservation_{0}, ..., \GPobservation_{N})$ from \rev{$N+1$} data points. Conditioned on this data set and a query input $\GPtestInput$, the expected value and variance of the $l$-th  GP is:
\begin{align}
  \GPpredictionMeanSingle{l}{\GPtestInput} & = \GPmean{\GPtestInput} + k(\GPtestInput,\GPinputStacked) (K + \noiseVar I)^{-1} \GPobservationStacked\\
  \GPpredictionVarSingle{l}{\GPtestInput} & = k(\GPtestInput, \GPtestInput) + \noiseVar - k(\GPtestInput, \GPinputStacked) (K + \noiseVar I)^{-1} k(\GPinputStacked,\GPtestInput)
\end{align}
where $K$ is a \rev{$(N+1) \times (N+1)$} kernel matrix with $K_{ij} = k(\GPinput_i, \GPinput_j)$.

In this work we use the squared exponential covariance as the kernel
\begin{equation}
    k(\testPointOne, \testPointTwo) = \sigma_f^2 \exp(-\frac{1}{2}(\testPointOne-\testPointTwo)^T\Sigma(\testPointOne-\testPointTwo)).
\end{equation}
where $\testPointOne$ and $\testPointTwo$ are any two inputs, $\Sigma$ is the lengthscale matrix which indicates the relevance between two data points and $\sigma_f^2$ is the signal variance which is a scale factor.
The hyperparameters ($\sigma, \sigma_f, \Sigma$) for a particular data set can be selected by maximum likelihood estimation, which maximizes the likelihood of the observed outputs given hyperparameters \cite{rasmussen2003gaussian}.




\subsection{Finding compensation signal using optimization}
From \eqref{eq:torque_definition2} it can be seen that if $\wrenchVecCmd\cSys{B}$ is set equal to the controller output $\wrenchVecDes\cSys{B}$, the actual resulting wrench $\wrenchVec\cSys{B}$ deviates from $\wrenchVecDes\cSys{B}$ by the error function $\unknownDynamicsFunc{\wrenchVecDes\cSys{B}}$.  

In order to achieve the desired wrench $\wrenchVecDes \cSys{B}$, we propose to add a compensation element $\dWrenchVecDes \in \R^6$ to the $\wrenchVecDes \cSys{B}$ when setting $\wrenchVecCmd \cSys{B}$, that is,
\begin{equation}\label{eq:torque_cmd_with_compensation}
  \dWrenchVecDes :=\wrenchVecCmd \cSys{B} - \wrenchVecDes \cSys{B} 
\end{equation}
A $\dWrenchVecDes$ is to be found such that when $\eqref{eq:torque_cmd_with_compensation}$ is substituted into $\eqref{eq:torque_definition2}$, the produced wrench $\wrenchVec \cSys{B}$ is equal to the desired wrench $\wrenchVecDes \cSys{B}$:
\begin{equation}
  \wrenchVecDes \cSys{B} \stackrel{!}{=} \wrenchVec \cSys{B}  = (\wrenchVecDes \cSys{B} + \dWrenchVecDes) + \unknownDynamicsFunc{\wrenchVecDes \cSys{B} + \dWrenchVecDes}.
\end{equation}


With  $\unknownDynamicsFunc{\cdot}$ modeled as GPs, we obtain
\begin{equation} \label{eq:fixedPointEq1}
  \dWrenchVecDes + \GPpredictionMean{\wrenchVecDes \cSys{B} + \dWrenchVecDes} = 0
\end{equation}
\rev{with $\GPpredictionMean{\cdot}$ being the mean vector function of the GPs.}

An online optimization problem is formulated to approximately solve \eqref{eq:fixedPointEq1}:
\begin{equation} \label{eq:cost_function}
  \underset{\dWrenchVecDes}{\text{minimize}} \quad \norm{\GPpredictionMean{\wrenchVecDes \cSys{B} + \dWrenchVecDes} + \dWrenchVecDes}_2^2 
\end{equation}
We optimize this in real time using gradient-based optimization.
If the optimal function value is \rev{smaller than 1e-4}, then we approximately find the compensation according to the mean function of the GP model. Otherwise, we compare the optimal value with the cost function value with $\dWrenchVecDes = 0$, and pick the \rev{corresponding signal} that yields a smaller cost.

\rev{\textit{Remark}: Formulating} the problem as an optimization problem has several advantages. Firstly, assuming the update rate is high enough, the previous solution is close to the current iteration solution, which is ideally suited as a warm start for the next iteration. Secondly, the search is local, therefore avoiding the jump between feasible solutions that are far away from each other. 
Due to the saturation function it is possible that the cost function has locally zero gradients. For this, a regularization term $||W_\mathrm{cmd}||_2^2$ can be added to the cost function.
Finally, the cost function is easy to manipulate and tune. We can add wrench command  constraints to the optimization problem or add costs that smoothen the adjacent wrench commands.

\subsection{Utilize posterior uncertainty prediction}
Since we are using a nonparametric method to learn the dynamics, this might lead to incorrect predictions for inputs that are far outside of the training data and thus destabilize the system. To address this issue we make use of the uncertainty information provided by the GP.
After we obtained the optimal solution $\dWrenchVecDes^*$ from the optimization, the compensation signal is filtered towards zero in case of high uncertainty of the prediction
\begin{equation}\label{eq:correction_filter}
    \dWrenchVecDes \leftarrow a \beta \dWrenchVecDesPrev + (1-\beta)\dWrenchVecDes^* 
\end{equation}
where $a \in (0,1)$, $\dWrenchVecDesPrev$ is the previously computed compensation signal,
\begin{equation}
\beta = \frac{1}{1+e^{-\kappa(\sigma - \sigma_{\mathrm{th}})}}
\end{equation}
where $\sigma = \max{\{\sigma_{l}(\wrenchVecDes \cSys{B} + \dWrenchVecDes^*), l=0,\dots,5\}}$ and with some constants $\sigma_{\mathrm{th}} \in \R$  and $\kappa \in \R$. 

For small prediction uncertainty $\beta$ is near zero and thus the compensation signal directly takes in the optimal value $\dWrenchVecDes^*$. 
For $\sigma$ larger than the threshold $\sigma_{\mathrm{th}}$, $\beta$ is near one.
$a$ determines how fast the compensation signal $\dWrenchVecDes$ should be filtered towards zero. $\kappa$ determines the rapidness of changing of $\beta$ near $\sigma_{\mathrm{th}}$.

Once the compensation signal is obtained, it is added to the output of the controller $\wrenchVecDes$. The resulting actuator command $\controlInputsCmd$ is obtained through the allocation mapping\textbf{} $\allocMapping(\cdot)$.

\subsection{Comparison with disturbance observer approaches}
\rev{An alternative approach is to employ an observer to estimate the unmodelled dynamics in real time and inform the controller to reject it. 
This approach is computationally efficient and has been shown to work well in practice. However, the approach introduces delay into tracking due to its nature of being a filter and furthermore requires expert knowledge to handcraft a disturbance model. 
The presented approach does not introduce delay and is less dependent on expert knowledge as it uses a data driven approach. The trade-off for this is the increased computational complexity and inability to compensate for unobserved disturbances. A good comparison of the properties of these two methods is presented in \cite{desaraju2017experience}.
}









%% file: images/control_strategy_v3.pdf_tex
\begingroup%
  \makeatletter%
  \providecommand\color[2][]{%
    \errmessage{(Inkscape) Color is used for the text in Inkscape, but the package 'color.sty' is not loaded}%
    \renewcommand\color[2][]{}%
  }%
  \providecommand\transparent[1]{%
    \errmessage{(Inkscape) Transparency is used (non-zero) for the text in Inkscape, but the package 'transparent.sty' is not loaded}%
    \renewcommand\transparent[1]{}%
  }%
  \providecommand\rotatebox[2]{#2}%
  \newcommand*\fsize{\dimexpr\f@size pt\relax}%
  \newcommand*\lineheight[1]{\fontsize{\fsize}{#1\fsize}\selectfont}%
  \ifx\svgwidth\undefined%
    \setlength{\unitlength}{498.06472122bp}%
    \ifx\svgscale\undefined%
      \relax%
    \else%
      \setlength{\unitlength}{\unitlength * \real{\svgscale}}%
    \fi%
  \else%
    \setlength{\unitlength}{\svgwidth}%
  \fi%
  \global\let\svgwidth\undefined%
  \global\let\svgscale\undefined%
  \makeatother%
  \begin{picture}(1,0.34842904)%
    \lineheight{1}%
    \setlength\tabcolsep{0pt}%
    \put(0,0){\includegraphics[width=\unitlength,page=1]{control_strategy_v3.pdf}}%
    \put(0.10521144,0.2939763){\color[rgb]{0,0,0}\makebox(0,0)[lt]{\lineheight{0.5}\smash{\begin{tabular}[t]{l}nominal \\controller\\\end{tabular}}}}%
    \put(0,0){\includegraphics[width=\unitlength,page=2]{control_strategy_v3.pdf}}%
    \put(0.83451027,0.28828329){\color[rgb]{0,0,0}\makebox(0,0)[lt]{\lineheight{1.25}\smash{\begin{tabular}[t]{l}control allocation \end{tabular}}}}%
    \put(0.60783884,0.25558781){\color[rgb]{0,0,0}\makebox(0,0)[lt]{\lineheight{1.25}\smash{\begin{tabular}[t]{l}+\end{tabular}}}}%
    \put(0,0){\includegraphics[width=\unitlength,page=3]{control_strategy_v3.pdf}}%
    \put(0.64046501,0.09027075){\color[rgb]{0,0,0}\makebox(0,0)[lt]{\lineheight{1.25}\smash{\begin{tabular}[t]{l}actuators\\\end{tabular}}}}%
    \put(0,0){\includegraphics[width=\unitlength,page=4]{control_strategy_v3.pdf}}%
    \put(0.30828406,0.09699229){\color[rgb]{0,0,0}\makebox(0,0)[lt]{\lineheight{1.25}\smash{\begin{tabular}[t]{l}system\\dynamics\end{tabular}}}}%
    \put(0,0){\includegraphics[width=\unitlength,page=5]{control_strategy_v3.pdf}}%
    \put(0.00849935,0.30417862){\color[rgb]{0,0,0}\makebox(0,0)[lt]{\lineheight{1.25}\smash{\begin{tabular}[t]{l}reference \end{tabular}}}}%
    \put(0.00216972,0.26546482){\color[rgb]{0,0,0}\makebox(0,0)[lt]{\lineheight{1.25}\smash{\begin{tabular}[t]{l}trajectory\end{tabular}}}}%
    \put(0.19596903,0.30607995){\color[rgb]{0,0,0}\makebox(0,0)[lt]{\lineheight{1.25}\smash{\begin{tabular}[t]{l}desired wrench\end{tabular}}}}%
    \put(0.01669293,0.21328636){\color[rgb]{0,0,0}\makebox(0,0)[lt]{\lineheight{1.25}\smash{\begin{tabular}[t]{l}state estimate\end{tabular}}}}%
    \put(0.78388265,0.06501109){\color[rgb]{0,0,0}\makebox(0,0)[lt]{\lineheight{1.25}\smash{\begin{tabular}[t]{l}actuator\\commands\end{tabular}}}}%
    \put(0,0){\includegraphics[width=\unitlength,page=6]{control_strategy_v3.pdf}}%
    \put(0.40548624,0.30716314){\color[rgb]{0,0,0}\makebox(0,0)[lt]{\lineheight{1.25}\smash{\begin{tabular}[t]{l}GP\end{tabular}}}}%
    \put(0.37035834,0.20600813){\color[rgb]{0,0,0}\makebox(0,0)[lt]{\lineheight{1.25}\smash{\begin{tabular}[t]{l}optimization\end{tabular}}}}%
    \put(0,0){\includegraphics[width=\unitlength,page=7]{control_strategy_v3.pdf}}%
    \put(0.6695771,0.30094712){\color[rgb]{0,0,0}\makebox(0,0)[lt]{\lineheight{1.25}\smash{\begin{tabular}[t]{l}command wrench\end{tabular}}}}%
    \put(0,0){\includegraphics[width=\unitlength,page=8]{control_strategy_v3.pdf}}%
    \put(0.45512508,0.1024193){\color[rgb]{0,0,0}\makebox(0,0)[lt]{\lineheight{1.25}\smash{\begin{tabular}[t]{l}generated wrench\end{tabular}}}}%
    \put(0,0){\includegraphics[width=\unitlength,page=9]{control_strategy_v3.pdf}}%
    \put(0.33556941,0.01515069){\color[rgb]{0,0,0}\makebox(0,0)[lt]{\lineheight{1.25}\smash{\begin{tabular}[t]{l}VoliroX\end{tabular}}}}%
    \put(0,0){\includegraphics[width=\unitlength,page=10]{control_strategy_v3.pdf}}%
    \put(0.53114067,0.33622555){\color[rgb]{0,0,0}\makebox(0,0)[lt]{\lineheight{1.25}\smash{\begin{tabular}[t]{l}compensation \\signal\end{tabular}}}}%
    \put(0,0){\includegraphics[width=\unitlength,page=11]{control_strategy_v3.pdf}}%
    \put(0.83898275,0.16327129){\color[rgb]{0,0,0}\makebox(0,0)[lt]{\lineheight{1.25}\smash{\begin{tabular}[t]{l}saturation function\end{tabular}}}}%
    \put(0,0){\includegraphics[width=\unitlength,page=12]{control_strategy_v3.pdf}}%
  \end{picture}%
\endgroup%

%% file: 06_Experiments.tex

\section{Experimental results} \label{sec:results}

The experiments are carried out at an indoor aerial vehicle testbed at Autonomous System Lab, ETH Zurich. A motion capture system provides pose estimates at 100 Hz.
The experimental vehicle platform VoliroX has a diameter of about 80 cm and a weight of 4 kg.
The vehicle is also equipped with an onboard NUC i7 computer, which runs computationally expensive modules, and a PixHawk flight controller that takes care of the low-level, low latency tasks. Using this configuration the vehicle is able to run all the necessary algorithms onboard.
For a more complete description see \cite{bodie2019omnidirectional}.

The proposed framework \rev{is implemented as a ROS node} and evaluated by the performance of attitude trajectory tracking \rev{both in simulation and on a real platform}.  \rev{In simulation, the framework is evaluated on a figure-8 trajectory with the vehicle tilting up to 63 degrees in the RotorS Gazebo simulator \cite{Furrer2016}.} For the real experiments, the vehicle follows a smooth feasible reference pitching trajectory from 0 to 60 degrees and back, while remaining stationary with zero roll and yaw. The maximal reference angular acceleration is 1 \si{\radian\per\square\second}. These trajectories were chosen as from experience the most prominent model-plant mismatch occurs on the torque level, especially when the produced force vector is close to the body $xy$ plane, corresponding to the vehicle attitude with high roll or pitch.


\subsection{GP model-plant mismatch modeling and learning}
This section shows the details of the model-plant mismatch using GP in real experiments and demonstrates its prediction performance on a validation trajectory.
As mentioned before, the most prominent model-plant mismatch occurs on the torque level, three single-output GPs are therefore used to model the torque model-plant-mismatch in the experiments and the force model-plant-mismatch is neglected.
For the GP regression, the inputs are the wrench command $\wrenchVecCmd \cSys{B}$ and the outputs are the components of $(\torqueVecMeas \cSys{B} - \torqueVecCmd\cSys{B})$, where $\torqueVecMeas \cSys{B}$ is \rev{obtained from a high-quality ADIS16448 IMU sensor \cite{adis}.} 
\rev{The training data is collected while the vehicle tracks a test trajectory with a nominal controller.
Sinusoidal excitation is added to the reference trajectory to collect diverse training data. This increases the probability of the learned model covering the output space during validation flight with the proposed strategy.}
Data points are subsampled from the experimental data offline using the $k$-medoids algorithm where \rev{the Euclidean squared distance between the inputs is} used as the distance metric.
Through empirical validation we found a hundred subsampled data points are sufficient for this trajectory.
The hyperparameters \rev{of the GPs} are fixed after \rev{being estimated} using the maximum likelihood methods, as we assume the model-plant-mismatch characteristics along the reference trajectory are repeatable. 
The GPs are implemented using GPy \cite{gpy2014}, the optimization uses L-BFGS implemented from nlopt \cite{nlopt}, and the $k$-medoids algorithm is taken from the PyClustering \cite{Novikov2019}. 
The learned model is then embedded into the online optimization framework, whose performance is then evaluated and presented in the following. 

\begin{figure}[h!]
\centering
\includegraphics{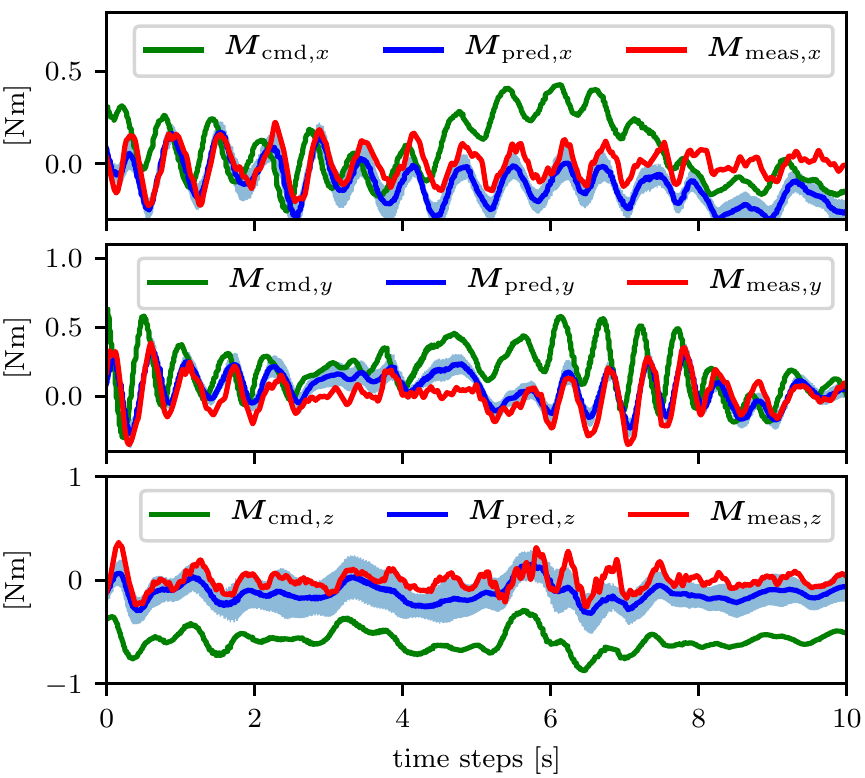}
\caption{Measurement, prediction (our model), and command of the torque vector (nominal model) of a 10 s trajectory. \rev{The 3 sigma confidence region of the GP prediction is shaded in blue.}}
\label{pic:experiment1}
\end{figure}

Fig.~\ref{pic:experiment1} shows the command torque $\torqueVecCmd$, the measured torque $\torqueVecMeas$, and the predicted torque $\torqueVecPred$ of a 10 seconds trajectory. $\torqueVecPred$ is $\GPpredictionMeanSingle{\torqueVec}{\wrenchVecCmd} + \torqueVecCmd$, where $\GPpredictionMeanSingle{\torqueVec}{\cdot}$ denotes the torque outputs of the GPs.
Table \ref{tab:prediction_performance} shows the mean and standard deviation of the absolute differences between the prediction and the measurement on the same trajectory (that is, $|\torqueVecCmd - \torqueVecMeas|$ and $|\torqueVecPred - \torqueVecMeas|$). 
It can be observed from Table \ref{tab:prediction_performance} that the prediction \rev{error} on the torque has been reduced by 62\%, 70\%, and 74\% on the body $x$, $y$ and $z$ axis, respectively. 
In addition, the mean of the prediction error using the learned model on the body $x$ and $y$ axis are comparable (0.08 and 0.067 Nm), while the error on the body $z$ axis is much larger (0.13 Nm). 
\rev{This is expected as the torque around the body $z$-axis is mainly generated by tilting the motor arms (especially during hover), while the torque around the body $x$- and $y$-axis is mainly induced by the sum of the propeller thrust times the motor arm length. While both servo dynamics and motor dynamics are neglected during the model learning, the prediction around the body $z$-axis is more affected since the servo dynamics are much slower than the motor dynamics.
}


\begin{table}[h!]
\renewcommand{\arraystretch}{1.3}
\caption{Prediction performance}
\label{tab:prediction_performance}
\centering
\begin{tabular}{c|c|c|c}
\hline
[Nm] & body $x$ error &  body $y$ error  &  body $z$ error   \\
\hline
nominal mean $\pm$ std & 0.21 $\pm$ 0.16 & 0.23 $\pm$ 0.16 & 0.50 $\pm$ 0.17 \\
\hline
learned mean $\pm$ std & 0.08 $\pm$ 0.06 & 0.07 $\pm$ 0.05 & 0.13 $\pm$ 0.09\\
\hline
\end{tabular}
\end{table}

\subsection{Trajectory tracking}
\subsubsection{Real experiments}
\begin{figure}[h!]
\centering
\includegraphics[width=0.8\linewidth]{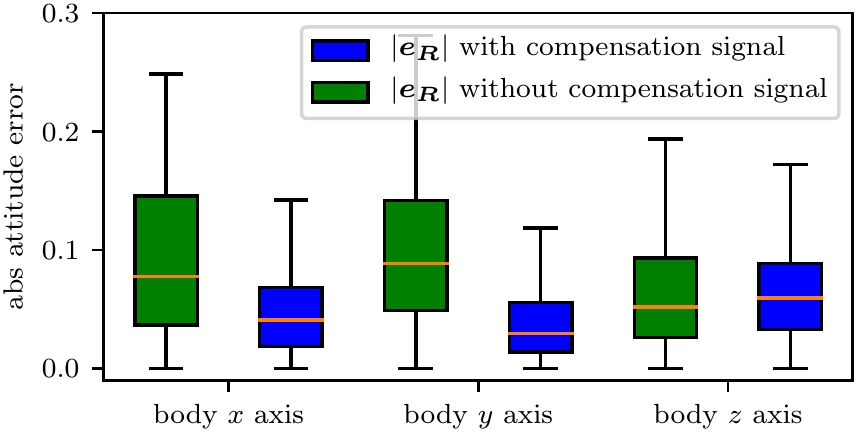}
\caption{A box plot comparison of the tracking performance for a pitching trajectory with and without the compensation signal. In both experiments, the same reference trajectory is executed consecutively 10 times. 
Note that by small angle approximation $\angErr$ corresponds to the Euler angles in radian.}
\label{pic:experiment2}
\end{figure}

The experiments demonstrated in this section compare the trajectory tracking performance between the cases with and without compensation signals.
Fig.~\ref{pic:experiment2} shows a box plot of the absolute values of the attitude tracking errors $|\angErr|$ from two experiments. In the first experiment the vehicle tracks the pitch trajectory 10 times consecutively using the control strategy without the compensation, whereas the second experiment repeats the same trajectory using the proposed framework. \rev{Running onboard a NUC i7 computer, it takes on average 7.5 ms and maximally 24.0 ms for solving one optimization problem in the proposed framework. } 

In both cases, an integral term is added to the attitude controller \eqref{eq:attitude_control}.
\rev{The gains of the nominal controller are tuned such that both approaches result in a stable flight in both experiments for the sake of comparison.
They are $(\dampingRatioPos$, $\naturalFreqPos) = (0.707, 3.5)$ and ($\dampingRatioAtt$, $\naturalFreqAtt$) = $(1.3, 3.5)$ and $(0.74, 3.5)$, where the former gain is tuned for roll and pitch axis and the latter gain tuned for the yaw axis. The integral attitude gain is 0.3.}

It can be seen that the medians of the absolute attitude tracking errors around the body $x$ and $y$ axis have reduced from 0.07, 0.087 to 0.04, 0.03, respectively, corresponding to a reduction of  43\% and 65\%. Their box ranges are also reduced from 0.114 and 0.10 to 0.050 to 0.040. 
On the other hand, there are no improvements shown on the body $z$ axis. The median increases from 0.051 to 0.058 and the box range decreases from 0.066 to 0.054. 
\rev{The tracking in body $z$-axis has not been improved mainly due to the following two reasons:
firstly, it can be seen in Fig.~\ref{pic:experiment1} that the model-plant mismatch is mostly a constant offset. In the nominal controller, this offset is already mitigated by the integrator term around the body-$z$ axis. 
Secondly, as mentioned in the previous section, the servo dynamics and the motor dynamics are not learned using the GP. 
To further improve the tracking performance in body $z$ axis, the input space of the GP could be augmented to include the past history of the wrench commands.}

\begin{figure}[h!]
\centering
\includegraphics[width=\linewidth]{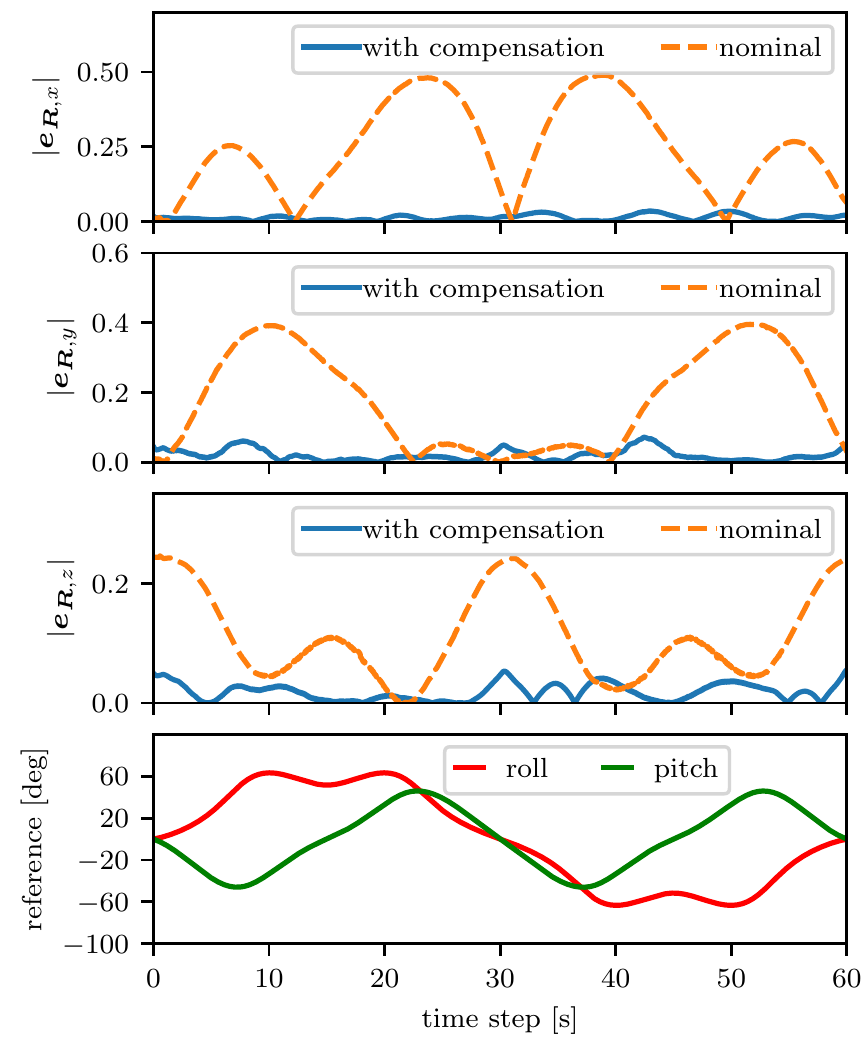}
\caption{\rev{VoliroX tracking a figure 8 reference trajectory with and without compensation signals in the simulation. Note that by small angle approximation $\angErr$ corresponds to the Euler angles in radian.}}
\label{pic:experiment3}
\end{figure}

\rev{\subsubsection{Simulation}
A simulation is conducted to demonstrate the generalizability of the learned model.
The training data is obtained by tilting the flying vehicle along various body axes up to 70 degrees  (with sinusoidal excitation around all three body axes). 200 data points are subsampled to train the GP model. A figure 8 trajectory with the vehicle tilting up to 63 degrees is then flown to evaluate the approach. A comparison of tracking performance with and without compensation signal is shown in \figref{pic:experiment3}.
The controller gains are again the same in both cases. Although this time the integral gains are set to zero.  
The tracking errors along 3 body axes have been reduced by $94.5\%, 87.9\%$, and $83.7 \%,$ respectively.
Similar to the real experiments, the key to having a good model is to excite all body axes during the fight. This ensures that the model can provide informative predictions for all required states.}

\subsection{Incorporation of prediction uncertainty}
\begin{figure}[h!]
\centering
\includegraphics[width=0.8\linewidth]{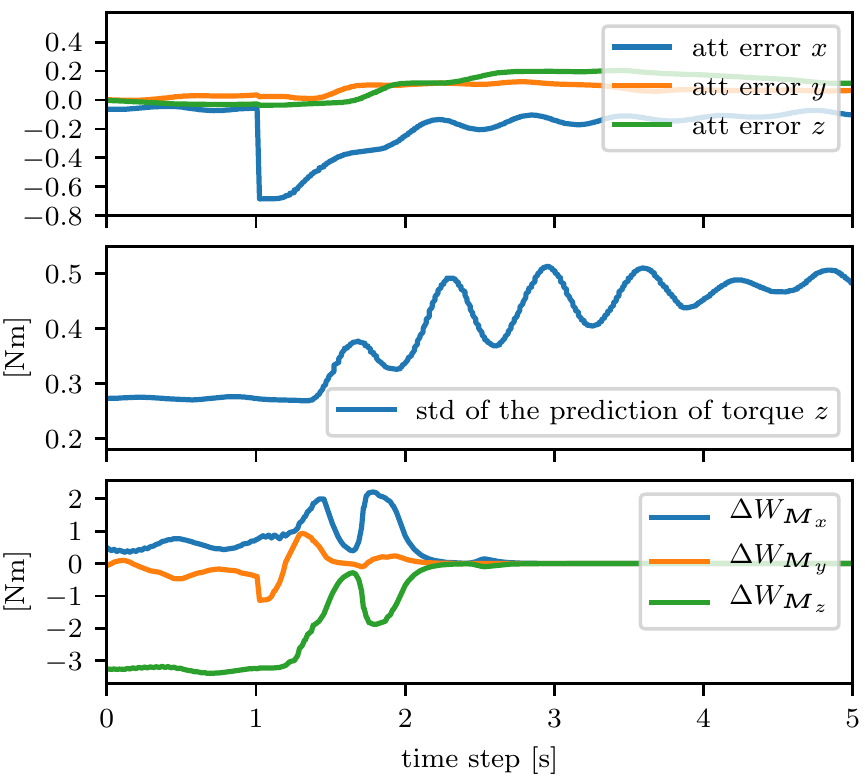}
\caption{This figure shows that for a step input (around 1.0 s) around body-$x$ direction. The desired wrench command is far away from the training data and GP prediction is uncertain at this query input. Therefore the compensation signals ($\dWrenchVecDes_{\torqueVec_x}, \dWrenchVecDes_{\torqueVec_y}, \dWrenchVecDes_{\torqueVec_z}$) get filtered towards zero according to \eqref{eq:correction_filter} and the vehicle remains stable. \rev{Note that by small angle approximation $\angErr$ corresponds to the Euler angles (roll, pitch, yaw from the reference frame to the body frame) in radian.}}
\label{pic:experiment4}
\end{figure}

The experiment presented in this section aims to demonstrate that even in face of highly uncertain predictions, the system is still robust thanks to the uncertainty provided by the model.
Two experiments using the learned model are conducted, one with the uncertainty check (that is, with performing \eqref{eq:correction_filter}) and the other one without. 
In both cases the vehicle in hover is given a step reference of 40 degrees roll.
Since the training data consists of smooth pitching trajectories, the wrench command for a step input in roll is far away from the training data.
Fig.~\ref{pic:experiment4} shows the experiment with the controller with uncertainty check.
The standard deviation of the prediction uncertainty in body $z$ direction increases from 0.27 to 0.5. 
The compensation signal is then filtered towards zero and the flying vehicle is able to remain stable.  
The parameters $\Sigma_{\mathrm{th}}, \kappa, a$ are heuristically chosen. 
In another experiment with the controller without the uncertainty check, the system becomes unstable due to the unreliable compensation signal\footnote{For a visualization of the behaviors, please see the attached video}.
\rev{In this case, it can be expected that the tracking performance of the approach with the uncertainty check cannot be better than the approach without compensation. GP will not predict well in region where no data is available.}

%% file: 07_Conclusion.tex

\section{Conclusion and outlook} \label{sec:conclusion}
This paper presented an approach that addresses the control challenges caused by the model-plant mismatches of an overactuated flying vehicle. 
Specifically, our approach learns \rev{a} well-defined forward model using \rev{a} GP and avoids the multi-valued mapping of the inverse model of the overactuated systems.
It finds a compensation signal through an optimization, which then cancels out the effect caused by the model-plant mismatch. 
In addition, the uncertainty prediction from the GP is exploited to prevent uncertain predictions from destabilizing the flying vehicle.  
Experiments on a real platform show that the proposed approach reduces the tracking error by 43\% and 65\% around body $x$- and $y$-axis and could prevent destabilizing the system in case of highly uncertain prediction of the model.



\section*{Acknowledgements}
The authors would like to thank Nicholas Lawrance for thoroughly going through the manuscript. This work was partly supported by funding from NCCR Digital Fabrications, NCCR Robotics and Armasuisse.